\documentclass[10pt,twocolumn,letterpaper]{article}

\usepackage[pagenumbers]{wacv} 

\usepackage{graphicx}
\usepackage{amsmath}
\usepackage{amssymb}
\usepackage{booktabs}

\usepackage[pagebackref,breaklinks,colorlinks]{hyperref}

\usepackage{orcidlink}

\usepackage{mathtools,lipsum}
\usepackage{multirow}
\usepackage{multicol}
\usepackage{caption}
\usepackage[inkscapelatex=false]{svg}
\usepackage{adjustbox}
\usepackage{indentfirst}
\usepackage{makecell}
\usepackage[T1]{fontenc}
\usepackage{tablefootnote}
\usepackage[flushleft]{threeparttable}
\usepackage[accsupp]{axessibility} 

\usepackage[capitalize]{cleveref}
\crefname{section}{Sec.}{Secs.}
\Crefname{section}{Section}{Sections}
\Crefname{table}{Table}{Tables}
\crefname{table}{Tab.}{Tabs.}

\def\wacvPaperID{267} 
\def\confName{WACV}
\def\confYear{2025}

\begin{document}

\title{PLReMix: Combating Noisy Labels with Pseudo-Label Relaxed Contrastive Representation Learning}

\author{Xiaoyu Liu, Beitong Zhou, Zuogong Yue, Cheng Cheng$^{*}$\\
{\tt\small Huazhong University of Science and Technology}\\
{\tt\small \{lxysl, zhoubt, z\_yue, c\_cheng\}@hust.edu.cn}
}
\maketitle

\begin{abstract}
Recently, the usage of Contrastive Representation Learning (CRL) as a pre-training technique improves the performance of learning with noisy labels (LNL) methods.
However, instead of pre-training, when trivially combining CRL loss with LNL methods as an end-to-end framework, the empirical experiments show severe degeneration of the performance.
We verify through experiments that this issue is caused by optimization conflicts of losses and propose an end-to-end \textbf{PLReMix} framework by introducing a Pseudo-Label Relaxed (PLR) contrastive loss.
This PLR loss constructs a reliable negative set of each sample by filtering out its inappropriate negative pairs, alleviating the loss conflicts by trivially combining these losses.
The proposed PLR loss is pluggable and we have integrated it into other LNL methods, observing their improved performance.
Furthermore, a two-dimensional Gaussian Mixture Model is adopted to distinguish clean and noisy samples by leveraging semantic information and model outputs simultaneously.
Experiments on multiple benchmark datasets demonstrate the effectiveness of the proposed method.
Code is available at \url{https://github.com/lxysl/PLReMix}.
\end{abstract}

\section{Introduction}
\label{sec:intro}

Deep neural networks (DNNs) lead to huge performance boosts in various classification tasks~\cite{krizhevsky2017imagenet, he2016deep, ren2015faster, ronneberger2015u}.
However, the large-scale, high-quality annotated datasets required by training DNNs are extremely expensive and time-consuming.
Many studies turn to collect data in less expensive but more efficient ways, such as querying search engines~\cite{li2017webvision, chen2015webly, cha2012social}
, crawling images with tags~\cite{mahajan2018exploring, liu2011noise, deng2020multi} or annotating with network outputs~\cite{kuznetsova2020open}.
Datasets collected in such alternative ways inevitably contain noisy labels.
These corrupted noisy labels severely interfere with the model's training process due to over-fitting, which leads to worse generalization performance~\cite{zhang2021understanding, arpit2017closer}.

\begin{figure}[t!]
  \centering
  \includegraphics[width=0.45\textwidth]{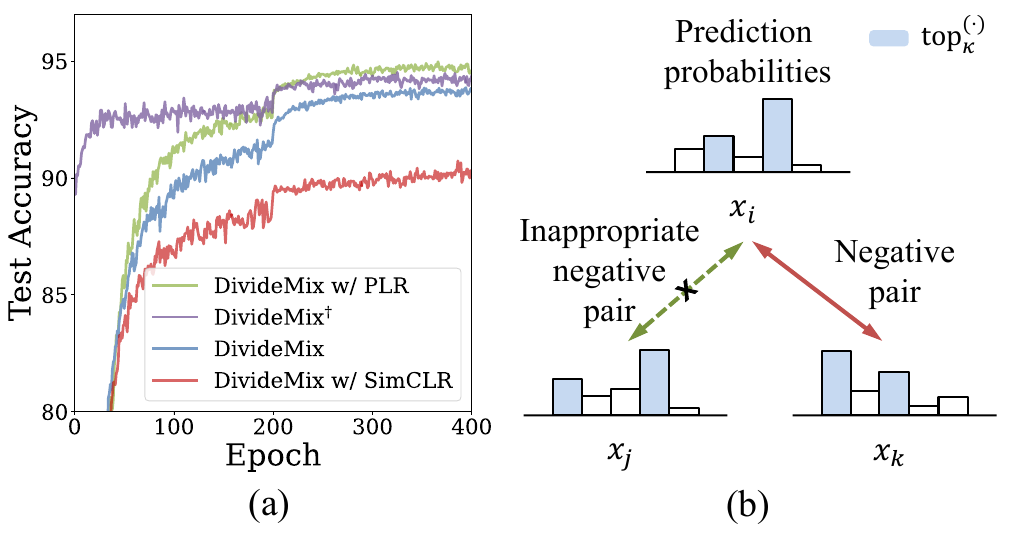}
  \caption{\textbf{Left}: Model performance suffers from trivially combining contrastive representation learning with supervised learning.
  The performance of DivideMix~\cite{li2019dividemix} (a supervised LNL method) boosts from using SimCLR~\cite{chen2020simple} pre-trained weights ($\mathrm{DivideMix}^\dag$), whereas suffers from being trivially combined with SimCLR contrastive loss ($\mathrm{DivideMix \ w/ \ SimCLR}$).
  While DivideMix with our proposed PLR contrastive loss ($\mathrm{DivideMix \ w/ \ PLR}$) achieves comparable results to $\mathrm{DivideMix}^\dag$.
  \textbf{Right}: Our proposed PLR contrastive loss selects a set of reliable negative pairs for each sample. The negative pair should have no overlap with the sample's prediction probabilities at $\mathrm{top}_\kappa$ indices ($\kappa = 2$ in Figure). As an instance, $x_k$ is a reliable negative pair of $x_i$ as $\mathrm{top}_2^{(i)} \cap \mathrm{top}_2^{(k)} = \emptyset$, while $x_j$ is an inappropriate negative pair of $x_i$ as $\mathrm{top}_2^{(i)} \cap \mathrm{top}_2^{(j)} \neq \emptyset$.}
  \label{fig:head}
  \vspace{-1em}
\end{figure}

Existing methods dealing with the so-called learning with noisy labels (LNL) problem can be mainly classified into two categories, namely label correction and sample selection.
Label correction methods leverage the model predictions to correct noisy labels~\cite{reed2014training, arazo2019unsupervised}.
Sample selection techniques attempt to divide samples with clean labels (usually with small losses) from noisy datasets~\cite{han2018co, jiang2018mentornet, li2019dividemix}.
Recent studies typically filter out the potentially noisy samples first, then apply label correction to those noisy samples, and finally train the model in a semi-supervised way~\cite{li2019dividemix}.

Most label correction and sample selection methods take advantage of \textit{early learning} phenomenon during the model training process~\cite{arpit2017closer}.
Nevertheless, the intrinsic semantic information in data inherently resistant to label noise memorization has been disregarded in these methods~\cite{wu2018unsupervised}.
The rapid development of self-supervised contrastive representation learning (CRL) methods~\cite{chen2020simple, he2020momentum} shows new possibilities for modeling data representations without the requirements of data labels, which is beneficial for the model to correct corrupted labels or filter out noisy samples in datasets.
Most recent studies either utilize the CRL to acquire robust pre-trained weights~\cite{sachdeva2023scanmix, zheltonozhskii2022contrast, ghosh2021contrastive} or directly add the CRL loss to the total loss accompanying the supervised loss~\cite{karim2022unicon, huang2023twin}.

While CRL helps models to learn representations without labels, a decrease in accuracy has been observed when it was trivially integrated with supervised losses (red curve in \cref{fig:head} left).
To address this issue, we propose the Pseudo-Label Relaxed (PLR) contrastive representation learning to avoid the conflict between supervised learning and CRL (\cref{fig:head} right).
Inappropriate negative pairs are removed by checking if their top $\kappa$ indices of prediction probabilities have an empty intersection.
This results in a reliable negative set and will construct a more effective semantic clustering than vanilla CRL.

In addition to improving model robustness to noisy labels by utilizing CRL, we also design a new technique to filter out noisy samples based on inconsistencies between labels and semantics.
Considering the semantic representation cluster formed by CRL, when the similarity between one sample's projection embedding with its label prototype is smaller than that between itself and its cluster prototype, it may be given the wrong label.
We adopt cross-entropy loss to formalize this sample selection method to obtain a consistent form with \textit{small loss selection} techniques.
Then, we dynamically fit a two-dimensional Gaussian Mixture Model (2d GMM) on the joint distribution of these two losses to divide clean and noisy samples.

Our main contributions are as follows:
\begin{itemize}
    \item We analyze the conflict between supervised learning and CRL, attributing it to objective inconsistency and gradient conflicts. A novel PLR loss is proposed to alleviate this issue, facilitating our end-to-end PLReMix framework to enhance the capacity for LNL problems. Our proposed PLR loss can be easily integrated into other LNL methods and boost their performance.
    \item We propose a sample selection method based on the inconsistency between semantic representation clusters and labels. A 2d GMM is utilized to filter out noisy samples, considering both the \textit{early learning} phenomenon of the model and intrinsic correlation in data.
    \item Extensive experiments on multiple benchmark datasets with varying types and label noise ratios demonstrate the effectiveness of our proposed method. We also conduct ablation studies and other analyses to verify the robustness of the components.
\end{itemize}

\section{Related Work}

\subsection{Learning with noisy labels}

\textbf{Label correction.}
Label correction methods attempt to use the model predictions to correct the noisy labels. To this end, during training, a noise transition matrix is estimated to correct prediction by transforming noisy labels to their corresponding latent ground truth~\cite{chen2015webly, sukhbaatar2015training}.
Huang \etal~\cite{huang2020self} update the noisy labels with model predictions through the exponential moving average.
P-correction~\cite{yi2019probabilistic} treats the latent ground truth labels as learnable parameters and utilizes back-propagation to update and correct noisy labels.

\textbf{Sample selection.}
Sample selection techniques attempt to filter the samples with clean labels from a noisy dataset.
Han \etal~\cite{han2018co} select samples with small losses from one network as clean ones to teach the other.
Dividemix~\cite{li2019dividemix} utilizes a two-component GMM to model the loss distribution to separate the dataset into a clean set and a noisy set.
Sukhbaatar \etal~\cite{wang2022promix} also consider the samples with consistent prediction and high confidence as clean samples.

Both label correction and sample selection methods benefit from the early learning phenomenon~\cite{li2020gradient, krause2016unreasonable, arpit2017closer}. That is to say, the weights of DNNs will not stray far from the initial weights in the early stage of training~\cite{li2020gradient}. Notably, several prior studies~\cite{krause2016unreasonable, arpit2017closer, song2019does, bai2021understanding} also observe the memorization effect in DNNs. This phenomenon demonstrates that DNNs tend to learn simple patterns and, over time, gradually overfit to more complex and noisy patterns.

\subsection{Contrastive representation learning}

\textbf{CRL in self-supervised learning.}
CRL, a representative self-supervised learning technique, holds great promise in learning label-irrelevant representations. Basically, the main idea of self-supervised learning is to design a pretext task that allows models to learn pre-trained weights. CRL learns label-irrelevant representations of the model that identify positive and negative examples corresponding to a single sample from a batch of data. In SimCLR~\cite{chen2020simple}, two random augmentations are applied to each sample to obtain pairs of positive samples, while the other augmented samples are considered negative examples.
A momentum encoder and a memory bank are adopted in MoCo~\cite{he2020momentum} to generate and retrieve negative features of samples, which alleviates the high demand for large batch sizes.

\begin{figure*}[htb!]
  \centering
  \includegraphics[width=0.95\textwidth, trim={0cm 0cm 0cm 0cm}, clip]{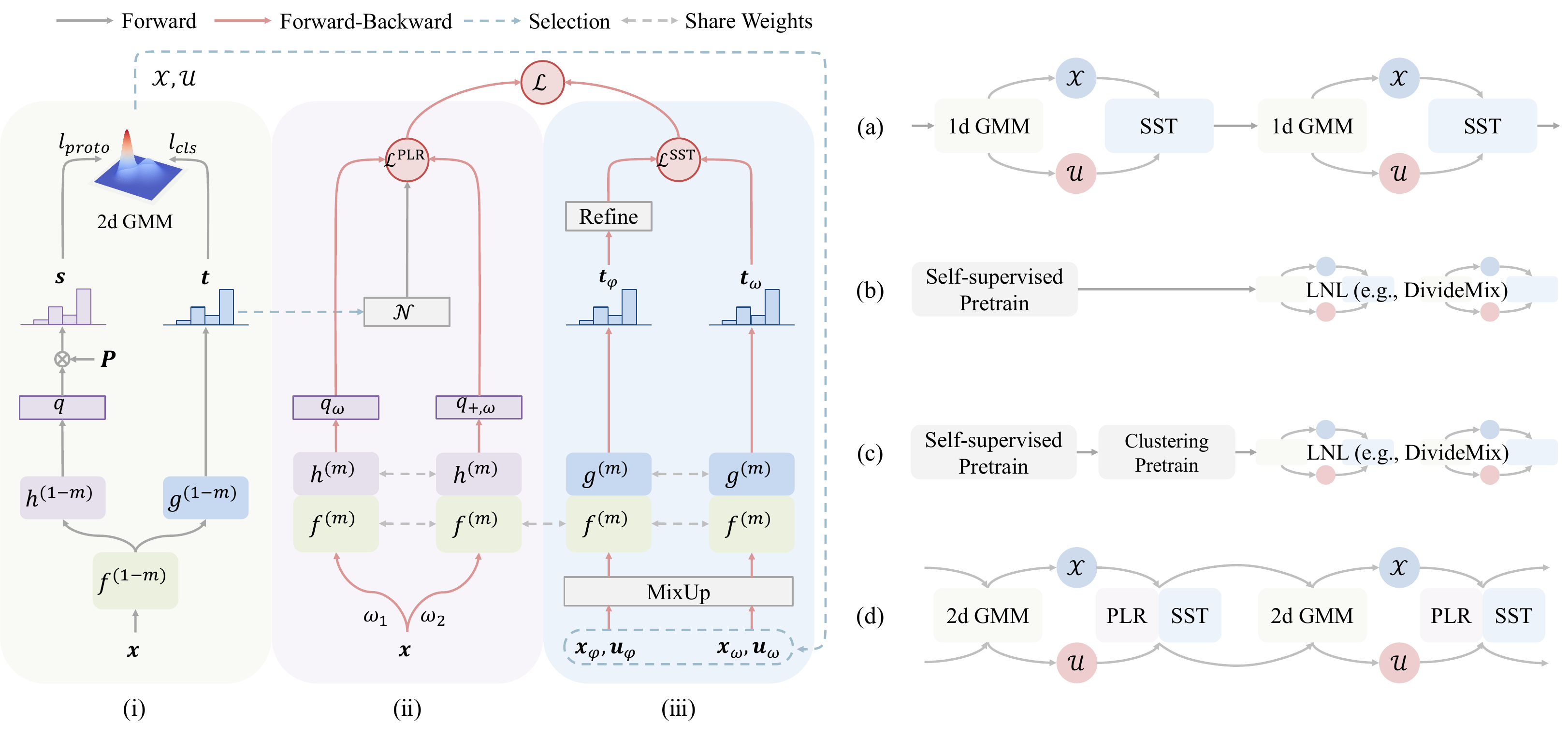}
  \caption{\textbf{Left}: Overview of the proposed method. (\romannumeral1) We perform joint sample selection on the first network to divide clean and noisy samples with 2d GMM. (\romannumeral2) We train the second network with proposed Pseudo-Label Relaxed contrastive loss $\mathcal{L}^{\mathrm{PLR}}$ with the constructed reliable negative set $\mathcal{N}$ alleviating the conflict between the supervised learning and CRL. (\romannumeral3) We utilize semi-supervised training for the second network on the clean and noisy samples divided by 2d GMM separately. The final learning objective $\mathcal{L}$ comprises two components, namely $\mathcal{L}^{\mathrm{PLR}}$ and  $\mathcal{L}^{\mathrm{SST}}$. \textbf{Right}: DivideMix and LNL frameworks utilizing CRL based on it. (a) DivideMix~\cite{li2019dividemix} iteratively performs sample selection (1d GMM) and semi-supervised training (SST). (b) The two-stage method \cite{zheltonozhskii2022contrast, ghosh2021contrastive, zhang2020decoupling} performs self-supervised pretraining before LNL. (c) The three-stage method~\cite{sachdeva2023scanmix} performs self-supervised pretraining and clustering pretraining before LNL. (d) By proposing PLR loss, our method unifies SST and CRL in an end-to-end PLReMix framework.}
  \label{fig:framework}
  \vspace{-1em}
\end{figure*}

\textbf{CRL in learning with noisy labels.}
As a label-free method, CRL is resistant to noisy labels and can be beneficial for the LNL problem.
Most of the recent studies follow a two or three-stage paradigm that uses CRL as a pretext task to obtain a pre-trained foundation model.
Zheltonozhskii \etal~\cite{zheltonozhskii2022contrast} and Ghosh \etal~\cite{ghosh2021contrastive} use CRL as a warmup method and outperform the fully-supervised pre-training on ImageNet.
ScanMix~\cite{sachdeva2023scanmix} uses the CRL following a deep clustering method to pre-train the model, then jointly train the network with semi-supervised loss and clustering loss.
Zhang \etal~\cite{zhang2020decoupling} utilizes CRL to pretrain the model, then trains a robust classification head on a fixed backbone, and finally trains in a semi-supervised way with a graph-structured regularization.
We compare these frameworks in \cref{fig:framework} right.

A few methods have been proposed to address the complexity associated with pretrain fine-tuning pipeline of CRL within the overall LNL framework.
Nevertheless, directly combining instance-wise CRL with supervised training objectives can interfere with the class-clustering ability, unlike forming clusters in a low-dimensional hypersphere surface in an unsupervised situation.
In MoPro~\cite{li2020mopro} and ProtoMix~\cite{li2021learning}, the prototypical CRL that enforces the sample embedding to be similar to its assigned prototypes is jointly used with standard instance-wise CRL.
MOIT~\cite{ortego2021multi} uses mixup interpolated supervised contrastive learning~\cite{khosla2020supervised} to improve model robustness to label noise.
UNICON~\cite{karim2022unicon} divides the dataset into clean and noisy sets, then trains in a semi-supervised manner with CRL loss applied on the noisy sets.
Sel-CL+~\cite{li2022selective} selects confident pairs by measuring the agreement between learned representations and labels to apply supervised contrastive learning to the LNL problem.
Kim \etal~\cite{kim2019nlnl} and Yan \etal~\cite{yan2022noise} propose negative learning approaches to alleviate the impact of noisy labels by teaching the model to learn this is \textit{not} something instead of this \textit{is} something.
Our method extends the self-supervised CRL by constructing a reliable negative set for each sample using pseudo-labels that the model predicts, thus alleviating the wrong contrast.

\section{Proposed Method}

\subsection{Overview}
The noisy dataset given in LNL is denoted as $\mathcal{D}=\{\boldsymbol{x}_i, y_i\}_{i=1}^N$, where $\boldsymbol{x}_i$ represents a sample and $y_i$ is the corresponding label over $C$ classes, $N$ is the number of samples in the dataset. The presence of noisy samples causes label differences between the annotated labels $\{y_i\}^N_{i=1}$ and the ground truth labels $\{\tilde{y}_i\}^N_{i=1}$. 
Our proposed framework iteratively trains two identical networks, each consisting of a convolutional backbone $\boldsymbol{f}(\cdot;\theta^{(m)})$, a classification head $\boldsymbol{g}(\cdot;\phi^{(m)})$, and a projection head $\boldsymbol{h}(\cdot;\psi^{(m)})$. $m \in \{0, 1\}$ denotes the network number.
Given a minibatch of $b$ samples, the output logits of input $\boldsymbol{x}=\{\boldsymbol{x}_i\}_{i=1}^{b}$ are denoted as $\boldsymbol{z}=\boldsymbol{g}(\boldsymbol{f}(\boldsymbol{x};\theta^{(m)});\phi^{(m)})$, the output prediction probabilities are denoted as $\boldsymbol{t}=\mathrm{softmax}(\boldsymbol{z})$, the projection embeddings are denoted as $\boldsymbol{q}=\boldsymbol{h}(\boldsymbol{f}(\boldsymbol{x};\theta^{(m)});\psi^{(m)})$.
We use a weak augmentation $\varphi$ and a strong augmentation $\omega$.
Class prototypes $\boldsymbol{P}=\{\boldsymbol{p}_k\}_{k=1}^C$, are maintained and updated every epoch to represent the feature prototypes of the $k$-th class, where $\boldsymbol{p}_k \in \mathbb{R}^{d}$, $d$ is the dimension of projection embedding.

For the initialization of class prototypes $\boldsymbol{P}$ and initial model convergence, we warm up the two networks for a few epochs with cross-entropy loss and standard CRL loss on the whole dataset.
At the end of the warmup, we initialize class prototypes by taking the average of feature embeddings with the same noisy labels.
As shown in \cref{fig:framework} left, the proposed end-to-end LNL framework performs the following two steps iteratively in the training process: (1) divide the whole dataset into a clean and a noisy set considering their losses by a novel two-component 2d GMM (\cref{GMM}) and (2) train the other network with proposed Pseudo-Label Relaxed contrastive loss $\mathcal{L}^{\mathrm{PLR}}$ on the representations with a set of reliable negative pairs $\mathcal{N}_i$ for each sample (\cref{PLR}), and semi-supervised loss $\mathcal{L}^{\mathrm{SST}}$ on the divided clean and noisy sets (\cref{SST}). $\mathcal{N}_i$ is denoted as $\mathcal{N}$ in \cref{fig:framework} left.

\subsection{Joint sample selection}\label{GMM}

\textbf{Sample selection}
\label{sec:sample_selection}
We compute two cross-entropy losses $l_{i, cls}$ and $l_{i, proto}$ to fit a two-component 2d GMM for sample selection. First, following the \textit{small loss selection} technique~\cite{han2018co}, we compute the cross-entropy classification loss $l_{i, cls} = -\log(\boldsymbol{t}_i^{y_i})$ to measure how well the model fits the pattern of a sample.
Then, we define the class prototypes as the center of projection embeddings with similar semantics. The similarity between each projection embedding $\boldsymbol{q}_i$ and all the class prototypes $\{\boldsymbol{p}_k\}^C_{k=1}$ represents the latent class probability distributions at the semantic level.
Following MoPro~\cite{li2020mopro}, we use softmax normalized cosine similarity to measure the similarity between $\boldsymbol{q}_i$ and $\boldsymbol{p}_k$:
\vspace{-0.6em}
\begin{equation}
    \boldsymbol{s}_i = \{s_i^k\}_{k=1}^C = \left\{ \frac{\exp{(\boldsymbol{q}_i \cdot \boldsymbol{p}_k / \tau_s)}}{\sum_{k=1}^C \exp{(\boldsymbol{q}_i \cdot \boldsymbol{p}_k / \tau_s)}} \right\}_{k=1}^C,
    \label{eq:similarity}
\vspace{-0.3em}
\end{equation}
where sharpen temperature $\tau_s$ is $0.1$ in this paper.
Projection embeddings for the samples with similar semantics tend to form clusters around corresponding prototypes under the effect of CRL.
In such a case, if a sample $\boldsymbol{x}_i$ is mislabeled to $y_i$, we suppose to have $s_i^{y_i} < s_i^{\tilde{y}_i}$, where $\tilde{y}_i$ is the latent ground truth class of $\boldsymbol{x}_i$.
Based on the above discussion, the distance between the distribution of $\boldsymbol{s}_i$ and OneHot$(y_i)$ is supposed to be smaller for clean samples than for noisy samples.
Thus, we can compute another cross-entropy loss $l_{i, proto} = -\log(\boldsymbol{s}_i^{y_i})$ to measure if the given label matches the semantic cluster.

After obtaining the $l_{i, cls}$ and the $l_{i, proto}$ for all samples,  a two-component 2d GMM on the distribution of $\boldsymbol{l} = \{\boldsymbol{l}_i\}_{i=1}^{N} = \{(l_{i, cls}, l_{i, proto})\}_{i=1}^N$ can be computed.
Suppose the two Gaussian components are $\boldsymbol{G_1} (\cdot; \boldsymbol{\mu}_1, \boldsymbol{\Sigma}_1)$ and $\boldsymbol{G_2} (\cdot; \boldsymbol{\mu}_2, \boldsymbol{\Sigma}_2)$, we obtain the probability $w_i$ of each sample to be clean from posterior probability as:
\vspace{-0.5em}
\begin{equation}
    w_i = p \left(\boldsymbol{G}_j \mid \boldsymbol{l}_i\right) \quad \mathrm{with} \quad 
 j={\mathop{\arg\min}\limits_{j \in \{1, 2\}} \left\Vert \boldsymbol{\mu}_j \right\Vert_2}. 
\vspace{-0.5em}
\end{equation}

Samples predicted have smaller $\left\Vert \boldsymbol{\mu}_j \right\Vert_2$ are considered as clean samples, which form the clean set denoted by $\mathcal{X}$.
While the remaining samples are considered as noisy and form the noisy set $\mathcal{U}$. 

\textbf{Momentum class prototypes}
We adopt class-aware self-adaptive sample selection~\cite{wang2022freematch} and exponential moving average (EMA) to update the class prototypes $\{\boldsymbol{p}_k\}^C_{k=1}$ while training.
To better estimate the latent ground truth label $\tilde{y}_i$, we generate a pseudo soft label $\hat{\boldsymbol{y}}_i = \alpha \boldsymbol{t}_i + (1-\alpha) \boldsymbol{s}_i$ for $\boldsymbol{x}_i$ considering both model predictions and semantic cluster.
Accordingly, a global threshold $\tau_g$ and class thresholds $\boldsymbol{\tau}_c = \left\{\tau_c^k\right\}_{k=1}^C$ are self-adaptively calculated and updated using EMA to filter confident samples:
\begin{equation}
\vspace{-0.5em}
\begin{aligned}
    \tau_g &\leftarrow \eta \tau_g + \left( 1-\eta \right) \frac{1}{N} \sum\nolimits_{i=1}^N \max\left( \hat{\boldsymbol{y}}_i \right), \\ 
    \tilde{\tau}_c^k &\leftarrow \eta \tilde{\tau}_c^k + \left( 1-\eta \right) \frac{1}{N} \sum\nolimits_{i=1}^N \hat{y}_i^k, \\
    \tau_c^k &= \left( \tilde{\tau}_c^k / \max \left\{ \tilde{\tau}_c^k \right\}_{k=1}^C \right) \cdot \tau_g. \\
\end{aligned}
\end{equation}
The selected high prediction confidence samples $\{\boldsymbol{x}_i, \hat{y}_i \mid \hat{y}_i = \max\left(\hat{\boldsymbol{y}}_i\right) > \tau_c^{\arg\max \left( \hat{\boldsymbol{y}}_i \right)}\}$ constitute the confident set denoted as $\mathcal{C}$.
With the estimated labels, we update the projection embedding $\boldsymbol{q}_i$ to the corresponding class prototype $\boldsymbol{p}_{\hat{y}_i}$ using EMA:
\begin{equation}
    \hspace{-0.5em}
    \boldsymbol{p}_{\hat{y}_i} \leftarrow \mathrm{Normalize} \left(\eta \boldsymbol{p}_{\hat{y}_i} + \left(1-\eta\right) \mathrm{Normalize}\left(\boldsymbol{q}_i\right)\right), \\
\end{equation}
for all ${\{\boldsymbol{x}_i, \hat{y}_i\}} \in \mathcal{C}$, where $\mathrm{Normalize}(\boldsymbol{x}) = \boldsymbol{x} / \left\Vert \boldsymbol{x} \right\Vert_2$ and the momentum coefficient $\eta$ is set to $0.99$.

\subsection{Pseudo-Label Relaxed contrastive representation learning}\label{PLR}
Typically, unsupervised CRL methods use two transforms of a single sample as a positive pair and combinations of transforms of other samples as negative pairs. We intend to design a loss function that will bring positive pairs together and push negative pairs apart.
Specifically, the vanilla InfoNCE loss for the embedding $\boldsymbol{q}_{i, \omega}$ of an augmented sample used in  CPC~\cite{oord2018representation} and SimCLR~\cite{chen2020simple} are defined as:
\begin{equation}
\vspace{-0.5em}
    - \sum_{i=1}^N \log \frac{\exp\left(\left\langle \boldsymbol{q}_{i, \omega}, \boldsymbol{q}_{+, \omega} \right\rangle / \tau \right)}{
    \splitdfrac{\exp\left(\left\langle \boldsymbol{q}_{i, \omega}, \boldsymbol{q}_{+, \omega} \right\rangle / \tau \right) + }
    {\sum_{\boldsymbol{q}_{j,\omega} \in \{\boldsymbol{q}_{-, \omega}\}}\exp\left(\left\langle \boldsymbol{q}_{i, \omega}, \boldsymbol{q}_{j, \omega} \right\rangle / \tau \right)}},
    \label{eq:info_nce}
\vspace{-0.3em}
\end{equation}
where $\tau$ is the sharpening temperature. $\{\boldsymbol{q}_{+,\omega}\}$ and $\{\boldsymbol{q}_{-,\omega}\}$ are the projection embeddings of augmented positive and negative samples, respectively, $\left\langle \boldsymbol{a}, \boldsymbol{b} \right\rangle$ represents cosine similarity $\mathrm{Normalize}\left( \boldsymbol{a} \cdot \boldsymbol{b} \right)$. We note that there exist some negative embeddings $\boldsymbol{q}_{j,\omega}\in \{\boldsymbol{q}_{-,\omega}\}$ whose ground truth labels are identical to the positive embedding $\boldsymbol{q}_{i,\omega}$. These samples will also be pushed away from $\boldsymbol{q}_{i,\omega}$ with the use of the term $\exp\left(\left\langle \boldsymbol{q}_{i, \omega}, \boldsymbol{q}_{j, \omega} \right\rangle / \tau \right)$ in Eq. \eqref{eq:info_nce}, which conflicts with the design purpose of other supervised losses used in LNL problems. 
Worse still, this effect is exacerbated when the $i$-th sample and $j$-th sample are more similar, resulting in a more dispersed cluster.

We proposed the PLR loss to address this issue, which is robust to both clean and noisy samples. A reliable negative set $\mathcal{N}_i$ is designed to remove the samples in $\{\boldsymbol{q}_{-,\omega}\}$ that share the same latent ground truth with $\boldsymbol{q}_{i,\omega}$.
To this end, given a prediction probability $\boldsymbol{t}_i=\left[t_{i,1}, t_{i,2},\ldots,t_{i,C}\right]$, we first identify the indexes of the top $\kappa$ digits of the prediction probability $\boldsymbol{t}_i$ of the $i$-th sample, which is denoted by $\mathrm{top}_\kappa^{(i)}$:
\vspace{-0.5em}
\begin{equation}
    \mathrm{top}_\kappa^{(i)} = \mathop{\arg\max}\limits_\kappa \{t_{i,\kappa}\}^C_{\kappa=1}.
\vspace{-0.5em}
\end{equation}
Thus, the reliable negative set of $\boldsymbol{x}_i$ can be obtained as:
\vspace{-0.5em}
\begin{equation}
    \left\{j \mid {\mathrm{top}_\kappa^{(i)}} \cap {\mathrm{top}_\kappa^{(j)}} = \emptyset, \forall j\right\}.
\vspace{-0.5em}
\end{equation}
During the early training stage, we additionally append the given label $y_i$ to the 
top$_\kappa^{(i)}$ for further conflict avoidance:
\vspace{-0.5em}
\begin{equation}
    \hspace{-0.6em}
    \mathcal{N}_i = \left\{j \mid \{\mathrm{top}_\kappa^{(i)} \cup y_i\} \cap \{\mathrm{top}_\kappa^{(j)} \cup y_j\} = \emptyset, \forall j \right\}.
\vspace{-0.5em}
\end{equation}
Finally, the proposed PLR loss is re-formulated based on the vanilla InfoNCE loss of Eq. \eqref{eq:info_nce} to the following expression:
\vspace{-1.4em}
\begin{equation}
    \hspace{-0.5em}
    \mathcal{L}^{\mathrm{PLR}} = - \sum_{i=1}^N \log \frac{\exp\left(\left\langle \boldsymbol{q}_{i, \omega}, \boldsymbol{q}_{+, \omega} \right\rangle / \tau \right)}
    {\splitdfrac{\exp\left(\left\langle \boldsymbol{q}_{i, \omega}, \boldsymbol{q}_{+, \omega} \right\rangle / \tau \right) + }{\sum_{j \in \mathcal{N}_i}\exp\left(\left\langle \boldsymbol{q}_{i, \omega}, \boldsymbol{q}_{j, \omega} \right\rangle / \tau \right)}}.
\vspace{-0.3em}
\end{equation}
We use two forms of PLR loss in this paper: the FlatNCE~\cite{chen2021simpler} formalized PLR loss is denoted as \textbf{FlatPLR}, while the InfoNCE formalized one is denoted just as \textbf{PLR}.
We do experiments on gradient analysis between $\mathcal{L}^{\mathrm{PLR}}$ and $\mathcal{L}^{\mathrm{CRL}}$ in \cref{gradient}, whose results show that $\mathcal{L}^{\mathrm{PLR}}$ has less gradient conflicts with the supervised loss.

\subsection{Semi-supervised training}\label{SST}

As described in \cref{sec:sample_selection}, according to the proposed sample selection technique, we divide the noisy dataset $\mathcal{D}$ into a labeled set $\mathcal{X}$ and an unlabeled set $\mathcal{U}$ using 2d GMM.
Then, we perform semi-supervised training on them following FixMatch~\cite{sohn2020fixmatch} and MixMatch~\cite{berthelot2019mixmatch}.

First, we generate pseudo targets $\overline{\boldsymbol{y}}_i$ for each sample using weak augmentations:
\vspace{-0.5em}
\begin{equation}
    \hspace{-1.3em}
    \resizebox{\hsize}{!}{$
    \begin{cases}
    \operatorname{Sharpen}\left(w_i \operatorname{OneHot}\left(y_i\right)+\frac{1-w_i}{\operatorname{card}(\varphi)} \sum_{\varphi} \boldsymbol{t}_{i, \varphi} ; T\right) , 
    \text { if } \boldsymbol{x}_i \in \mathcal{X}, \\ \operatorname{Sharpen}\left(\frac{1}{\operatorname{card}(\varphi) \operatorname{card}(m)} \sum_{\varphi, m} \boldsymbol{t}_{i, \varphi}^m ; T\right) , 
    \text { if } \boldsymbol{x}_i \in \mathcal{U},
    \end{cases}
    $}
\end{equation}
and
\vspace{-0.7em}
\begin{equation}
\hspace{-0.7em}
\vspace{-0.5em}
    \mathrm{Sharpen}(\bar{\boldsymbol{y}}_i; T) = {\bar{\boldsymbol{y}}_i^{c \frac{1}{T}}} / \sum_{c=1}^C{\bar{\boldsymbol{y}}_i^{c \frac{1}{T}}}, \ \mathrm{for} \ c=1,2,...,C,
\end{equation}
where ${\mathrm{card}}(\varphi)$ and ${\mathrm{card}}(m)$ represent the numbers of augmentations and models, respectively, and the sharpening temperature $T$ is $0.5$.
Then we do MixUp~\cite{zhang2018mixup} on the strong augmented samples $\boldsymbol{x}_{i, \omega}$ and pseudo targets $\bar{\boldsymbol{y}}_i$:
\vspace{-0.5em}
\begin{equation}
\vspace{-0.5em}
\begin{aligned}
    \boldsymbol{x}_{i, \omega}^\prime &= \lambda \boldsymbol{x}_{i, \omega} + (1-\lambda) \boldsymbol{x}_{r(i), \omega}, \\
    \bar{\boldsymbol{y}}_i^\prime &= \lambda \bar{\boldsymbol{y}}_i + (1-\lambda) \bar{\boldsymbol{y}}_{r(i)}, 
\end{aligned}
\end{equation}
where $\lambda \sim \mathrm{Beta}(\beta, \beta)$, $r(i)$ represents the random permutation of indices $i$.
Finally, the semi-supervised loss $\mathcal{L}^{\mathrm{SST}}$ consists of three parts: a cross-entropy loss on the labeled set, a mean squared error on the unlabeled set, and a regularization term $\mathcal{L}_{reg}$ used in~\cite{tanaka2018joint} and~\cite{arazo2019unsupervised}:
\vspace{-1em}
\begin{multline}
    \mathcal{L}^{\mathrm{SST}} = - \frac{1}{|\mathcal{X}|} \sum_{\boldsymbol{x}_i \in \mathcal{X}} \sum_{c=1}^C\bar{\boldsymbol{y}}_i^{\prime c}\log(\boldsymbol{t}_{i, \omega}^{\prime c}) + \\
    \lambda_{\mathcal{U}}\frac{1}{|\mathcal{U}|} \sum_{\boldsymbol{x}_i \in \mathcal{U}} \left\Vert \bar{\boldsymbol{y}}_i^\prime - \boldsymbol{t}_{i, \omega}^\prime \right\Vert_2^2 + \mathcal{L}_{\mathrm{reg}},
\vspace{-1em}
\end{multline}
where $\lambda_{\mathcal{U}}$ is a coefficient for tuning.
The final objective objective $\mathcal{L}$ is to minimize the following function:
\vspace{-0.5em}
\begin{equation}
    \mathcal{L} = \mathcal{L}^{\mathrm{SST}} + \lambda_i \mathcal{L}^{\mathrm{PLR}},
\vspace{-0.5em}
\end{equation}
where $\lambda_i$ is a tunable hyperparameter.

\section{Experiments}

\setlength{\tabcolsep}{4pt}
\begin{table*}[thb!]
\scriptsize
\centering
\caption{Experimental and comparison results on CIFAR-10 and CIFAR-100 datasets with different noise ratios  (test accuracy \%). \textbf{w/o 2d GMM} denotes the experiment where replace the 2d GMM with 1d GMM.}
\label{tb:results}
\begin{adjustbox}{width=0.8\textwidth}
\begin{threeparttable}
\begin{tabular}{@{}rccccccccccc@{}}
\toprule
\multirow{3}{*}{\textbf{Method}}                                                                                  & \multicolumn{1}{|c|}{\textbf{Dataset}}                          & \multicolumn{5}{c|}{\textbf{CIFAR-10}}                       & \multicolumn{5}{c}{\textbf{CIFAR-100}}   \\ \cmidrule(l){2-12}
                                                                                                         & \multicolumn{1}{|c|}{\textbf{Mode}} & \multicolumn{4}{c|}{\textbf{Sym}} & \multicolumn{1}{c|}{\textbf{Asym}} & \multicolumn{4}{c}{\textbf{Sym}} & \multicolumn{1}{|c}{\textbf{Asym}}  \\ \cmidrule(l){2-12}
                                                                                                         & \multicolumn{1}{|c|}{$\boldsymbol{r}$} & \textbf{20\%} & \textbf{50\%} & \textbf{80\%} & \textbf{90\%} & \multicolumn{1}{|c|}{\textbf{40\%}} & \textbf{20\%} & \textbf{50\%} & \textbf{80\%} & \textbf{90\%} & \multicolumn{1}{|c}{\textbf{40\%}} \\ \midrule
\multicolumn{1}{r|}{\multirow{2}{*}{Cross-entropy}} & \multicolumn{1}{c|}{Best}        & 86.8 & 79.4 & 62.9 & 42.7 & \multicolumn{1}{|c|}{85.0}     & 62.0 & 46.7 & 19.9 & 10.1 & \multicolumn{1}{|c}{44.5} \\
\multicolumn{1}{c|}{}                                                                                    & \multicolumn{1}{c|}{Last}        & 82.7 & 57.9 & 26.1 & 16.8 & \multicolumn{1}{|c|}{72.3}     & 61.8 & 37.3 & 8.8 & 3.5 & \multicolumn{1}{|c}{-} \\ \midrule
\multicolumn{1}{r|}{\multirow{2}{*}{MixUp (18')~\cite{zhang2018mixup}}}                                                       & \multicolumn{1}{c|}{Best}        & 95.6 & 87.1 & 71.6 & 52.2 & \multicolumn{1}{|c|}{-}     &  67.8 & 57.3 & 30.8 & 14.6 & \multicolumn{1}{|c}{48.1} \\
\multicolumn{1}{c|}{}                                                                                    & \multicolumn{1}{c|}{Last}        & 92.3 & 77.3 & 46.7 & 43.9 & \multicolumn{1}{|c|}{77.7}     & 66.0 & 46.6 & 17.6 & 8.1 & \multicolumn{1}{|c}{-} \\ \midrule
\multicolumn{1}{r|}{\multirow{2}{*}{Co-teaching+ (19')~\cite{yu2019does}}}                                                       & \multicolumn{1}{c|}{Best}        & 89.5 & 85.7 & 67.4 & 47.9 & \multicolumn{1}{|c|}{-}     &  65.6 & 51.8 & 27.9 & 13.7 & \multicolumn{1}{|c}{-} \\
\multicolumn{1}{c|}{}                                                                                    & \multicolumn{1}{c|}{Last}        & 88.2 & 84.1 & 45.5 & 30.1 & \multicolumn{1}{|c|}{-}     & 64.1 & 45.3 & 15.5 & 8.8 & \multicolumn{1}{|c}{-} \\ \midrule
\multicolumn{1}{r|}{\multirow{2}{*}{DivideMix (19')~\cite{li2019dividemix}}}                                                       & \multicolumn{1}{c|}{Best}        & 96.1 & 94.6 & 93.2 & 76.0 & \multicolumn{1}{|c|}{93.4}     & 77.3 & 74.6 & 60.2 & 31.5 & \multicolumn{1}{|c}{55.1} \\
\multicolumn{1}{c|}{}                                                                                    & \multicolumn{1}{c|}{Last}        & 95.7 & 94.4 & 92.9 & 75.4 & \multicolumn{1}{|c|}{92.1}     & 76.9 & 74.2 & 59.6 & 31.0 & \multicolumn{1}{|c}{-} \\ \midrule
\multicolumn{1}{r|}{{ELR+ (20')~\cite{liu2020early}}}                                                       & \multicolumn{1}{c|}{Best}        & 95.8 & 94.8 & 93.3 & 78.7 & \multicolumn{1}{|c|}{93.3}     &  77.6 & 73.6 & 60.8 & 33.4 & \multicolumn{1}{|c}{73.2} \\ \midrule
\multicolumn{1}{r|}{{MOIT+ (21')~\cite{ortego2021multi}}}                                                      & \multicolumn{1}{c|}{Best}        & 94.1 & 91.8 & 81.1 & 74.7 &  \multicolumn{1}{|c|}{93.3}     & 75.9 & 70.6 & 47.6 & 41.8 & \multicolumn{1}{|c}{74.0} \\ \midrule
\multicolumn{1}{r|}{Sel-CL+$^\dag$ (22')~\cite{li2022selective}} & \multicolumn{1}{c|}{Best}        & 95.5 & 93.9 & 89.2 & 81.9 & \multicolumn{1}{|c|}{93.4}     & 76.5 & 72.4 & 59.6 & 48.8 & \multicolumn{1}{|c}{72.7} \\ \midrule
\multicolumn{1}{r|}{{UNICON (22')~\cite{karim2022unicon}}}                                                       & \multicolumn{1}{c|}{Best}        & 96.0 & 95.6 & 93.9 & 90.8 & \multicolumn{1}{|c|}{94.1}     &  \textbf{78.9} & 77.6 & 63.9 & 44.8 & \multicolumn{1}{|c}{\textbf{74.8}} \\ \midrule
\multicolumn{1}{r|}{\multirow{2}{*}{ScanMix$^\dag$ (23')~\cite{sachdeva2023scanmix}}}                                                       & \multicolumn{1}{c|}{Best}        & 96.0 & 94.5 & 93.5 & 91.0 & \multicolumn{1}{|c|}{93.7}     & 77.0 & 75.7 & 66.0 & \textbf{58.5} & \multicolumn{1}{|c}{-} \\
\multicolumn{1}{c|}{}                                                                                    & \multicolumn{1}{c|}{Last}        & 95.7 & 93.9 & 92.6 & 90.3 & \multicolumn{1}{|c|}{93.4}     & 76.0 & 75.4 & 65.0 & \textbf{58.2} & \multicolumn{1}{|c}{-} \\ \midrule
\multicolumn{1}{r|}{{TCL (23')~\cite{huang2023twin}}}                                                       & \multicolumn{1}{c|}{Best}        & 95.0 & 93.9 & 92.5 & 89.4 & \multicolumn{1}{|c|}{92.7}     &  78.0 & 73.3 & 65.0 & 54.5 & \multicolumn{1}{|c}{-} \\ \midrule
\multicolumn{1}{r|}{\multirow{2}{*}{\textbf{\textbf{\makecell[c]{(Flat, w/o 2d GMM) \\ PLReMix}}}}}                                                        & \multicolumn{1}{c|}{Best}        & 96.38 & 95.63 & 94.89 & 92.36 &  \multicolumn{1}{|c|}{\textbf{95.42}}    & 78.08 & 76.94 & \textbf{69.27} & 48.97 & \multicolumn{1}{|c}{66.49} \\
\multicolumn{1}{c|}{}                                                                                    & \multicolumn{1}{c|}{Last}        & 96.18 & 95.31 & 94.70 & 92.16 & \multicolumn{1}{|c|}{\textbf{94.98}}     & 77.66 & 76.70 & \textbf{68.96} & 48.66 & \multicolumn{1}{|c}{61.25} \\ \midrule
\multicolumn{1}{r|}{\multirow{2}{*}{\textbf{\textbf{\makecell[c]{(Flat) \\ PLReMix}}}}}                                                        & \multicolumn{1}{c|}{Best}        & \textbf{96.63} & \textbf{95.71} & \textbf{95.08} & \textbf{92.74} &  \multicolumn{1}{|c|}{95.11}    & 77.95 & \textbf{77.78} & 68.76 & 50.17 & \multicolumn{1}{|c}{64.89} \\
\multicolumn{1}{c|}{}                                                                                    & \multicolumn{1}{c|}{Last}        & \textbf{96.46} & \textbf{95.36} & \textbf{94.84} & \textbf{92.43} & \multicolumn{1}{|c|}{94.72}     & 77.78 & \textbf{77.31} & 68.41 & 49.44 & \multicolumn{1}{|c}{62.67} \\ \bottomrule
\end{tabular}
\begin{tablenotes}
    \footnotesize
    \item{$^\dag$} denotes using pre-trained model weights.
\end{tablenotes}
\end{threeparttable}
\end{adjustbox}
\vspace{1em}
\end{table*}

\subsection{Experimental settings}
\textbf{CIFAR-10/100}
The CIFAR-10/100 dataset~\cite{krizhevsky2009learning} comprises 50K training images and 10K test images.
We adopt symmetric and asymmetric synthetic label noise models as used in previous studies~\cite{tanaka2018joint}.
The symmetric noise setting replaces $r\%$ of labels from one class uniformly with all possible classes, while asymmetric replaces them from one class only with the most similar class.

We use the PreAct ResNet-18~\cite{he2016identity} network on CIFAR 10/100 and train it using an SGD optimizer.
To avoid conflict early and gain robust representation later, we reduce $\kappa$ from $3$ to $2$ to $1$ at the epochs of 40 and 70.
In this work, we use AutoAug~\cite{cubuk2019autoaugment} as a strong augmentation operation, which is demonstrated to be useful in the LNL problem~\cite{nishi2021augmentation}.
We use FlatPLR for CIFAR experiments to gain a greater thrust between samples and their negative pairs with a relatively small batch size.
For more training details, please refer to the Appendix.

\textbf{Tiny-ImageNet}
Tiny-ImageNet~\cite{le2015tiny} consists of 200 classes, each containing 500 smaller-resolution images from ImageNet.
We train a PreAct ResNet 18~\cite{he2016identity} network with a batch size of 128 and a learning rate of 0.01.

\textbf{Clothing1M}
Clothing1M~\cite{xiao2015learning}, a large-scale real-world dataset with label noise, contains 1M clothing images from 14 classes.
We sample 64K images in total from each class~\cite{li2019dividemix}.
We train a ResNet50 network with weights pre-trained on ImageNet~\cite{he2016deep} for 100 epochs.

\textbf{WebVision}
WebVision dataset~\cite{li2017webvision} contains 2.4 M images crawled from Flickr and Google.
All of the images are categorized into 1,000 classes, which is the same as ImageNet ILSVRC12~\cite{olga15ilsvrc}.
We use the first 50 classes of the Google image subset to train an InceptionResnet V2~\cite{szegedy2017inception} from scratch.
Multi-crop augmentation strategy~\cite{caron2020unsupervised} is exploited to obtain more robust representations in rather shorter epochs and smaller batch size so that contrastive loss can be jointly trained with the semi-supervised loss.
See the Appendix for more details about multi-crop.

\begin{table}[t!]
    \vspace{-2em}
    \caption{Experimental and comparison results on Tiny-ImageNet dataset (test accuracy \%).}
    \vspace{-2em}
    \begin{center}
    \label{tb:tiny}
    \begin{adjustbox}{width=0.45\textwidth}
    \begin{tabular}{@{}r|c|cccc|c@{}}
    \toprule
    \multirow{2}{*}{\textbf{Method}} & \multirow{2}{*}{$\boldsymbol{r}$} & \multicolumn{4}{c|}{\textbf{Sym}} & \textbf{Asym} \\ \cmidrule(l){3-7}
     & & 0 & 20\% & 50\% & 80\% & 45\% \\ \midrule
    \multicolumn{1}{r|}{\multirow{2}{*}{Cross-entropy}} & Best & 57.4 & 35.8 & 19.8 & - & 26.3 \\
    \multicolumn{1}{c|}{} & Last & 56.7 & 35.6 & 19.6 & - & 26.2 \\ \midrule
    \multicolumn{1}{r|}{\multirow{2}{*}{Co-teaching+ (19')~\cite{yu2019does}}} & Best & 52.4 & 48.2 & 41.8 & - & 26.9 \\
    \multicolumn{1}{c|}{} & Last & 52.1 & 47.7 & 41.2 & - & 26.5 \\ \midrule
    \multicolumn{1}{r|}{\multirow{2}{*}{M-correction (19')~\cite{arazo2019unsupervised}}} & Best & 57.7 & 57.2 & 51.6 & - & 24.8 \\
    \multicolumn{1}{c|}{} & Last & 57.2 & 56.6 & 51.3 & - & 24.1 \\ \midrule
    \multicolumn{1}{r|}{\multirow{2}{*}{UNICON (22')~\cite{karim2022unicon}}} & Best & \textbf{63.1} & 59.2 & 52.7 & - & - \\
    \multicolumn{1}{c|}{} & Last & \textbf{62.7} & 58.4 & 52.4 & - & - \\ \midrule
    \multicolumn{1}{r|}{\multirow{2}{*}{\textbf{\makecell[c]{(Flat) \\ PLReMix}}}} & Best & 62.93 & \textbf{60.70} & \textbf{54.94} & \textbf{37.47} & \textbf{33.97} \\
    \multicolumn{1}{c|}{} & Last & 62.46 & \textbf{60.39} & \textbf{54.31} & \textbf{36.43} & \textbf{33.25} \\ \bottomrule
    \end{tabular}
    \end{adjustbox}
    \end{center}
    \vspace{-2.5em}
\end{table}

\begin{table}[t!]
    \vspace{-2em}
    \caption{Experimental and comparison results on Clothing1M dataset (test accuracy \%).}
    \vspace{-2em}
    \begin{center}
    \label{tb:clothing1m}
    \begin{adjustbox}{width=0.23\textwidth}
    \begin{threeparttable}
    \begin{tabular}{@{}r|c@{}}
        \toprule
        \textbf{Method} & \textbf{Accuracy} \\ \midrule
        Cross-Entropy & 69.21 \\ 
        Joint-Optim (18')~\cite{tanaka2018joint} & 72.16 \\ 
        P-correction (19')~\cite{yi2019probabilistic} & 73.49 \\ 
        DivideMix (19')~\cite{li2019dividemix} & 74.76 \\
        ELR+ (20')~\cite{liu2020early} & 74.81 \\
        C2D$^\dag$ (22')~\cite{zheltonozhskii2022contrast} & 74.30 \\
        UNICON (22')~\cite{karim2022unicon} & \textbf{74.95} \\
        ScanMix$^\dag$ (23')~\cite{sachdeva2023scanmix} & 74.35 \\
        TCL (23')~\cite{huang2023twin} & 74.8 \\ \midrule
        \textbf{PLReMix} & \textbf{74.85} \\
        \bottomrule
    \end{tabular}
    \begin{tablenotes}
        \footnotesize
        \item{$^\dag$} denotes using pre-trained model weights.
    \end{tablenotes}
    \end{threeparttable}
    \end{adjustbox}
    \end{center}
    \vspace{-3em}
\end{table}

\subsection{Experimental results}
The proposed end-to-end PLReMix framework is compared with other LNL methods on different datasets using the same network architecture.
The experimental results on CIFAR-10/100 are shown in \cref{tb:results}.
We report the best test accuracy over all epochs and the average test accuracy of the last 10 epochs as Best and Last, respectively.
Considering the intrinsic semantic features by utilizing PLR loss and 2d GMM, our proposed method outperforms state-of-the-art (SOTA) results in most experimental settings, especially on data with high noise ratios.
We attribute our inferior results to SOTA on CIFAR-100 with 40\% asymmetric noise to worse initialization of class prototypes across too many classes, and the absence of a uniform sample selection technique proved to be useful~\cite{karim2022unicon}.
However, our method still outperforms the baseline method~\cite{li2019dividemix} a lot.
In the second-to-last line, we substitute the 2d GMM with the previous 1d GMM (indicated as \textbf{w/o 2d GMM}), demonstrating that it yields inferior results compared to 2d GMM.

The experimental results on Tiny-ImageNet are shown in \cref{tb:tiny}.
Our algorithm achieves SOTA performance on most noise ratios.
The results of the real-world noisy dataset Clothing1M are shown in \cref{tb:clothing1m}, and the results of the WebVision dataset are listed in \cref{tb:webvision}.
Our algorithm achieves performance comparable to SOTA.
It is notable that C2D~\cite{zheltonozhskii2022contrast} and ScanMix~\cite{sachdeva2023scanmix} use pre-trained weights in a two/three-stage manner.
As a comparison, our proposed method trains the model end-to-end with randomly initialized weights.

\setlength{\tabcolsep}{4pt}
\begin{table}[t!]
\scriptsize
\centering
\caption{Experimental and comparison results on WebVision and ILSVRC12 datasets (test accuracy \%). Models are trained on the WebVision dataset and tested on both the WebVision and ILSVRC12 datasets. Top-1 and Top-5 test accuracy are reported.}
\vspace{-1em}
\label{tb:webvision}
\begin{adjustbox}{width=0.45\textwidth}
\begin{threeparttable}
\begin{tabular}{@{}r|c|cc|cc@{}}
\toprule
\textbf{Dataset}    & \multirow{2}{*}{\textbf{Architecture}} & \multicolumn{2}{c|}{\textbf{WebVision}} & \multicolumn{2}{c}{\textbf{ILSVRC12}} \\ \cmidrule(l){1-1} \cmidrule(l){3-6}
\textbf{Method}     & & \textbf{Top-1} & \textbf{Top-5} & \textbf{Top-1} & \textbf{Top-5} \\ \midrule
Co-Teaching (18')~\cite{han2018co}        & Inception V2 & 63.6 & 85.2 & 61.5 & 84.7 \\
DivideMix (19')~\cite{li2019dividemix}        & Inception V2 & 77.3 & 91.6 & 75.2 & 90.8 \\
ELR+ (20')~\cite{liu2020early}        & Inception V2 & 77.8 & 91.7 & 70.3 & 89.8 \\
UNICON (22')~\cite{karim2022unicon}        & Inception V2 & 77.6 & 93.4 & 75.3 & \textbf{93.7} \\
TCL (23')~\cite{huang2023twin}        & Inception V2 & 79.1 & 92.3 & 75.4 & 92.4 \\
ScanMix$^\dag$ (23')~\cite{sachdeva2023scanmix}        & Inception V2 & 80.0 & 93.0 & 75.8 & 92.6 \\ \midrule
\textbf{PLReMix}       & Inception V2 & \textbf{81.49} & \textbf{93.79} & \textbf{77.75} & 93.08 \\ \bottomrule
\end{tabular}
\begin{tablenotes}
    \footnotesize
    \item{$^\dag$} denotes using pre-trained model weights.
\end{tablenotes}
\end{threeparttable}
\end{adjustbox}
\vspace{-2em}
\end{table}

\subsection{PLR as a pluggable component}

We do extra experiments on the CIFAR-10 dataset with other different LNL algorithms to demonstrate the effectiveness and generalization of our proposed PLR loss.
We integrate our proposed PLR loss into Co-teaching+~\cite{yu2019does} and JoCoR~\cite{wei2020combating} and test the model performance under different noise ratios.
We use PreAct ResNet18 network, and the minor modification to the original algorithms is just training a projection head using our proposed PLR loss on top of the model backbone.
As a comparison, we utilize the vanilla SimCLR to train the projection head.
The experimental results are listed in \cref{tb:scalability}.
As can be seen, training with our proposed PLR increases the model performance compared to the baseline methods, while training with vanilla SimCLR tends to make the model performance corrupted.
These results align with our analysis of the gradient conflicts in \cref{gradient}.

\setlength{\tabcolsep}{4pt}
\begin{table}[t]
    \scriptsize
    \centering
    \caption{Experimental results of combining different CRL methods with LNL algorithms (test accuracy \%).}
    \vspace{-1em}
    \label{tb:scalability}
    \begin{adjustbox}{width=0.45\textwidth}
    \begin{tabular}{@{}c|c|cc|c@{}}
        \toprule
        \multirow{2}{*}{\textbf{Method}} & \multirow{2}{*}{\textbf{CRL}} & \multicolumn{3}{c}{\textbf{CIFAR-10}} \\ \cmidrule(l){3-5}
         & & \textbf{Sym-20}\% & \textbf{Sym-50}\% & \textbf{Asym-40}\% \\ \midrule
        \multirow{3}{*}{Co-teaching+~\cite{yu2019does}} & - & 88.73 & 77.52 & 68.64 \\
         & SimCLR & 88.62 & 80.24 & 68.86 \\
         & PLR & \textbf{90.28} & \textbf{85.68} & \textbf{70.92}\\ \midrule
        \multirow{3}{*}{JoCoR~\cite{wei2020combating}} & - & 88.13 & 80.94 & 75.48 \\ 
         & SimCLR & 84.76 & 78.81 & 78.56 \\ 
         & PLR & \textbf{89.34} & \textbf{83.46} & \textbf{81.46} \\
        \bottomrule
    \end{tabular}
    \end{adjustbox}
\vspace{-1em}
\end{table}

\subsection{Analysis}\label{analysis}

\textbf{Multi-task gradients.}
\label{gradient}
Given two losses $\mathcal{L}^{\left( 1 \right)}$ and $\mathcal{L}^{\left( 2 \right)}$ and their gradients to model parameters $\boldsymbol{g}^{\left( 1 \right)}$ and $\boldsymbol{g}^{\left( 2 \right)}$, we define the intensity of gradient entanglement between them as $\mathcal{E}^{\left( 1 \right)}_{\left( 2 \right)} = \boldsymbol{g}^{\left( 1 \right)} \cdot \boldsymbol{g}^{\left( 2 \right)} / {\left\Vert \boldsymbol{g}^{\left( 2 \right)} \right\Vert}_2^2$, the ratio of gradient magnitude between them as $\mathcal{R}^{\left( 1 \right)}_{\left( 2 \right)} = {\left\Vert \boldsymbol{g}^{\left( 1 \right)} \right\Vert}_2 / {\left\Vert \boldsymbol{g}^{\left( 2 \right)} \right\Vert}_2$.

\begin{figure}[t!]
  \centering
  \includegraphics[width=0.45\textwidth]{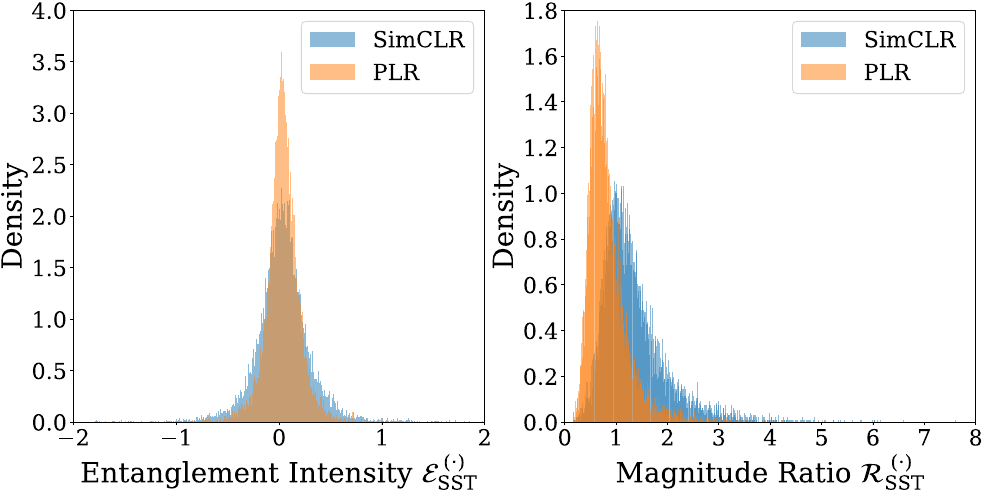}
  \caption{Gradient conflicts of two contrastive representation learning methods (vanilla SimCLR and our proposed PLR) when they are jointly trained with semi-supervised learning in the presence of noisy labels.}
  \label{fig:gradient}
  \vspace{-2em}
\end{figure}

Conflicting gradients and a large difference in gradient magnitudes often lead to optimization challenges in high-dimensional neural network multi-task learning~\cite{yu2020gradient}.
Similar issues are witnessed in our experiments.
Here, $\mathcal{L}^{\mathrm{SimCLR}}$ or $\mathcal{L}^{\mathrm{PLR}}$ is treated as $\mathcal{L}^{\left( 1 \right)}$, $\mathcal{L}^{\mathrm{SST}}$ is treated as $\mathcal{L}^{\left( 2 \right)}$.
We show the distribution of $\mathcal{E}^{\mathrm{SimCLR}}_{\mathrm{SST}}$ and $\mathcal{E}^{\mathrm{PLR}}_{\mathrm{SST}}$ (Left), $\mathcal{R}^{\mathrm{SimCLR}}_{\mathrm{SST}}$ and $\mathcal{R}^{\mathrm{PLR}}_{\mathrm{SST}}$ (Right) when training models on CIFAR-10 dataset with 80\% noise ratio at epoch 30 in \cref{fig:gradient}.

The entanglement intensity between SimCLR and SST shows more negative values, and their magnitude ratio has more values greater than 1.
These indicate that there are more gradient conflicts between $\mathcal{L}^{\mathrm{SST}}$ and $\mathcal{L}^{\mathrm{SimCLR}}$, which result in an optimization challenge and performance decrease shown in \cref{fig:head} left red line.
Our proposed PLR loss mitigates the gradient conflicts with SST, which manifests as more orthogonal gradients and smaller gradient magnitude ratios.
As a result, it improves the performance when integrated with SST to combat noisy labels, as shown in \cref{fig:head} left green line.

\textbf{Ablation study.}
We perform thorough ablation studies on our method.
The results of ablation studies are shown in \cref{tb:ablation}.
We test the effectiveness of FlatNCE by swapping it with (non-Flat) PLR, which is formalized by InfoNCE loss.
Despite using InfoNCE instead of FlatNCE, our experiments still show the robustness of the proposed PLR loss.
Then, we substitute the PLR loss with vanilla SimCLR~\cite{chen2020simple}.
Supervised contrastive loss (SCL)~\cite{khosla2020supervised} treats samples with the same labels as positive pairs, and others are negative pairs.
It can be a substitute for our proposed PLR loss in our framework when it considers samples with the same predicted pseudo labels as positive pairs.
We do experiments to compare SCL and PLR in our proposed framework.
The results show that our proposed PLR outperforms SCL in most noise ratios except on the CIFAR-100 dataset with a high noise ratio.
Experimental results about replacing 2d GMM with 1d GMM are shown in \cref{tb:results}, which confirm that semantic information helps differentiate clean and noisy samples on the loss distribution.

\setlength{\tabcolsep}{4pt}
\begin{table}[t!]
\scriptsize
\centering
\caption{Ablation studies on CIFAR-10 and CIFAR-100 datasets with different training settings (test accuracy \%).}
\vspace{-1em}
\label{tb:ablation}
\begin{adjustbox}{width=0.45\textwidth}
\begin{tabular}{@{}cccccccc@{}}
\toprule
\multirow{2}{*}{\textbf{Method}}                                                                                  & \multicolumn{1}{|c|}{\textbf{Dataset}}                          & \multicolumn{3}{c|}{\textbf{CIFAR-10}}                       & \multicolumn{3}{c}{\textbf{CIFAR-100}}   \\ \cmidrule(l){2-8} 
                                                                                                         & \multicolumn{1}{|c|}{$\boldsymbol{r}$} & \textbf{50\%} & \textbf{80\%} & \multicolumn{1}{c|}{\textbf{90\%}} & \textbf{50\%} & \textbf{80\%} & \textbf{90\%} \\ \midrule
\multicolumn{1}{c|}{\multirow{2}{*}{\textbf{\makecell[c]{(Flat) \\ PLReMix}}}}                                                        & \multicolumn{1}{c|}{Best}        & \textbf{95.71} & \textbf{95.08} &  \multicolumn{1}{c|}{\textbf{92.74}}    & \textbf{77.78} & \textbf{68.76} & 50.17 \\
\multicolumn{1}{c|}{}                                                                                    & \multicolumn{1}{c|}{Last}        & \textbf{95.36} & \textbf{94.84} & \multicolumn{1}{c|}{\textbf{92.43}}     & \textbf{77.31} & \textbf{68.41} & 49.44 \\ \midrule
\multicolumn{1}{c|}{\multirow{2}{*}{\textbf{PLReMix}}}                                                       & \multicolumn{1}{c|}{Best}        & \textbf{95.71} & 93.58 & \multicolumn{1}{c|}{91.71}     &  75.64 & 66.98 & 50.78 \\
\multicolumn{1}{c|}{}                                                                                    & \multicolumn{1}{c|}{Last}        & 95.31 & 93.39 & \multicolumn{1}{c|}{91.46}     & 75.39 & 66.76 & 50.54 \\ \midrule
\multicolumn{1}{c|}{\multirow{2}{*}{\makecell[c]{vanilla \\ SimCLR}}} & \multicolumn{1}{c|}{Best}        & 93.44 & 90.44 & \multicolumn{1}{c|}{86.80}     & 73.10 & 63.50 & 47.76 \\
\multicolumn{1}{c|}{}                                                                                    & \multicolumn{1}{c|}{Last}        & 93.23 & 90.28 & \multicolumn{1}{c|}{86.51}     & 72.32 & 63.02 & 46.81 \\ \midrule
\multicolumn{1}{c|}{\multirow{2}{*}{SCL}}                                                       & \multicolumn{1}{c|}{Best}        & 94.37 & 93.41 & \multicolumn{1}{c|}{91.24}     & 72.55 & 68.43 & \textbf{54.07} \\
\multicolumn{1}{c|}{}                                                                                    & \multicolumn{1}{c|}{Last}        & 94.08 & 93.11 & \multicolumn{1}{c|}{91.00}     & 72.18 & 68.03 & \textbf{53.72} \\ \bottomrule
\end{tabular}
\end{adjustbox}
\vspace{-1em}
\end{table}

\setlength{\tabcolsep}{4pt}
\begin{table}[t!]
\scriptsize
\centering
\caption{Ablation study results when using different $\kappa$ or only given labels on CIFAR-10 dataset with 80\% symmetric label noise.}
\vspace{-1em}
\label{tab:ablation_k}
\begin{adjustbox}{width=0.35\textwidth}
\begin{tabular}{@{}c|c|c|c|c|c@{}}
\toprule
 & $\boldsymbol{\kappa=3}$ & $\boldsymbol{\kappa=2}$ & $\boldsymbol{\kappa=1}$ & \textbf{only} $\boldsymbol{y}$ & \textbf{PLReMix} \\ \midrule
Best & 91.83 & 93.93 & 93.74 & 94.74 & \textbf{95.08} \\
Last & 91.29 & 92.81 & 93.60 & 94.58 & \textbf{94.84} \\
\bottomrule
\end{tabular}
\end{adjustbox}
\vspace{-2em}
\end{table}

We conduct ablation experiments on the hyperparameter $\kappa$.
Intuitively, larger values of $\kappa$ avoid conflict between PLR and supervised loss more effectively but result in less robust representation due to insufficient negative pairs.
To address this issue, we decrease $\kappa$ from 3 to 2 to 1 after training a certain number of epochs.
The decrease epochs are not carefully tuned and are set roughly evenly.
In this ablation study, we fix the value of $\kappa$ as 3, 2, or 1 during different individual training procedures.
In addition, we test the performance only using the given labels (denoted by only $y$), which means $\mathcal{N}_i = \left\{ j \mid y_i \cap y_j = \emptyset, \forall j \right\}$.
The experiments are conducted on the CIFAR-10 dataset with 80\% symmetric noise, and the results are listed in \cref{tab:ablation_k}.
The experimental results verify our analysis and show the effectiveness of the proposed decrease setting.

\begin{figure}[t]
    \begin{minipage}[t]{\linewidth} 
        \centering
        \includegraphics[width=0.9\textwidth]{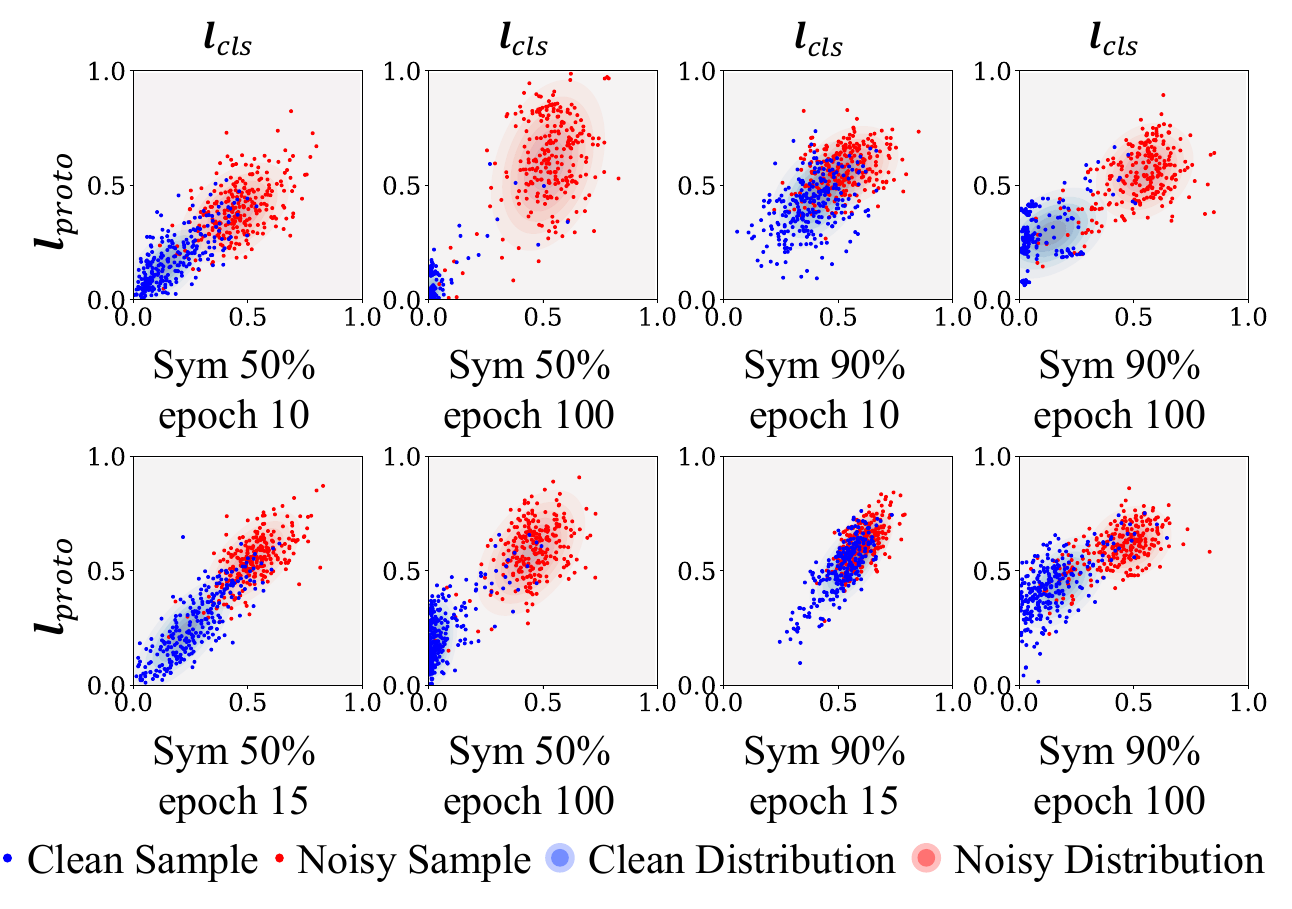}
        \vspace{-0.5em}
        \caption{Visualization of the 2d GMM fitted on the normalized loss distribution $\left\{\left(\boldsymbol{l}_{cls}, \boldsymbol{l}_{proto}\right)\right\}$. The first and second rows are obtained from experiments on the CIFAR-10 and CIFAR-100, respectively.}
        \label{fig:gmm}
    \end{minipage} %
    \hspace*{\fill}
    \begin{minipage}[t]{\linewidth} 
        \centering
        \includegraphics[width=0.9\textwidth]{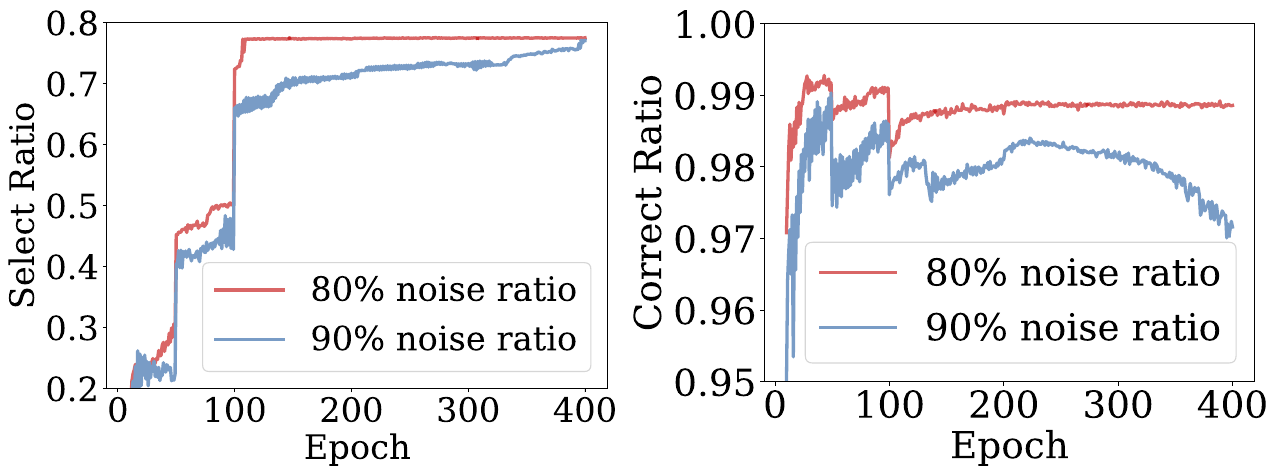}
        \vspace{-0.5em}
        \caption{Negative sample selection when training with FlatPLR. \textbf{Left}: Negative pairs selected ratio. \textbf{Right}: Correct negative pairs selected ratio.}
        \label{fig:negative}
    \end{minipage} %
    \vspace{-2em}
\end{figure}

\textbf{Visualization.}
We visualize the 2d GMM after 10 epochs of warming up and 100 epochs of training in \cref{fig:gmm}.
We plot the losses of 512 randomly selected samples, with half from the clean set and the other half from the noisy set.
As illustrated, our proposed 2d GMM effectively fits the distribution of the clean and noisy samples.

\textbf{Correct ratio of negative pairs.} We visualize the correct ratio of selected negative pairs in \cref{fig:negative} for a better understanding of the proposed PLR loss.
The select ratio is the proportion of selected negative sample pairs out of all possible negative sample pairs. The correct ratio represents the proportion of correctly identified negative pairs among all selected pairs.
We observed that the PLR loss tends to select a limited number of negative pairs at first to ensure high precision.
As $\kappa$ gradually decreases, PLR loss will rapidly increase the number of negative sample pairs selected while maintaining precision.

\section{Conclusion and Future Work}

In this work, we propose an end-to-end framework for solving the LNL problem by leveraging CRL.
We analyze conflicts between CRL learning objectives and supervised ones and propose PLR loss to address this issue in a simple yet effective way.
For further utilizing label-irrelevant intrinsic semantic information, we propose a joint sample selection technique, which expands previously widely used 1d GMM to 2d.
Extensive experiments conducted on benchmark datasets show performance improvements of our method over existing methods.
Our end-to-end framework shows significant improvement without any complicated pretrain fine-tuning pipeline.
Future research should also explore other forms of PLR loss, such as MoCo-based.

\section*{Acknowledgement}

This work is supported by the National Natural Science Foundation of China [Grant numbers 52205520, 62373160, 62273053].

{\small
\bibliographystyle{ieee_fullname}
\bibliography{egbib}
}

\end{document}